\documentclass[preprint,3p]{elsarticle}      

\makeatletter
\def\ps@pprintTitle{%
 \let\@oddhead\@empty
 \let\@evenhead\@empty
 \def\@oddfoot{\centerline{\thepage}}%
 \let\@evenfoot\@oddfoot}
\makeatother

\usepackage{amsmath}
\usepackage{natbib}
\usepackage{amssymb}
\usepackage{graphicx}
\usepackage{dcolumn}
\usepackage{bm}
\usepackage{mathrsfs}
\usepackage{color}
\usepackage{psfrag}
\usepackage{url}
\usepackage{soul}
\usepackage{float}  
\usepackage{cancel}

\usepackage[normalem]{ulem}
\usepackage{color}

\begin{document}

\title{Interpretable deep-learning models to help achieve the Sustainable Development Goals}

\author[focal1,focal2]{Ricardo Vinuesa\corref{cor1}}
\ead{rvinuesa@mech.kth.se}
\cortext[cor1]{Corresponding Author}
\author[focal3]{Beril Sirmacek}
\ead{b.sirmacek@saxion.nl}

\address[focal1]{FLOW, Engineering Mechanics, KTH Royal Institute of Technology, Stockholm, Sweden}
\address[focal2]{AI Sustainability Center, Stockholm, Sweden}
\address[focal3]{Smart Cities, School of Creative Technology, Saxion University of Applied Sciences, Enschede, The Netherlands}

\date{\today}


\begin{abstract}

We discuss our insights into interpretable artificial-intelligence (AI) models, and how they are essential in the context of developing ethical AI systems, as well as data-driven solutions compliant with the Sustainable Development Goals (SDGs). We highlight the potential of extracting truly-interpretable models from deep-learning methods, for instance via symbolic models obtained through inductive biases, to ensure a sustainable development of AI. 

\end{abstract}

\maketitle


Recent interest in artificial-intelligence (AI) methods has led to their application in a progressively wider range of applications, and their impact in our daily lives should not be underestimated. Despite the significant benefits of AI technology to improve our well-being~\cite{vinuesa_et_al_2020}, there are a number of areas where AI can hinder the achievement of a sustainable future~\cite{naude_vinuesa}. This dilemma has been recognized by Hilbert~\cite{hilbert}, and one of the main limitations of current AI technology in this context is the lack of interpretability of these models. The implications of this are nicely articulated by Rudin~\cite{rudin}, who claims that AI models need to be inherently interpretable and not mere ``black boxes'' or provide limited and shallow explainability. Another important example of interpretability of data-driven models can be found in the context of digital contact tracing for handling the coronavirus disease-19 (COVID-19) pandemic: there should be a right to contest the decisions made by the algorithm, and interpretability would be essential in this~\cite{vinuesa_covid}.

AI algorithms have the potential to support the achievement of the Sustainable Development Goals (SDGs) of the United Nations (UN)~\cite{vinuesa_et_al_2020}. For instance, Jean~{\it et al.}~\cite{jean_et_al} proposed a pioneering method to identify and track regions of poverty using satellite images, via convolutional neural networks (CNNs), a study with important implications for SDG 1 (on no poverty). However, we discuss below that it is essential to add interpretability to this type of model in order to develop efficient strategies to tackle this SDG. A similar observation can be made regarding SDG 13 (on climate action), where Chantry~{\it et al.}~\cite{Chantry} stated that the complexity of the AI models, combined with the numerous unknown parameters, make it extremely challenging to create a robust climate model, especially for forecasting applications. Another challenge for providing a robust model was of course the difficulty of generalizing one climate model to all the areas of Earth where measurements are conducted. On the other hand, Huntingford~{\it et al.}~\cite{Huntingford} highlighted the fact that AI models are typically black boxes, a fact that complicates identifying the origin of errors, the relative importance of the various parameters, and generally complicates climate-change research. They also argue that black-box AI models also generate uncertainty regarding their acceptability by government or the general public when they produce suggestions and/or make decisions. One well-known example is the medical application \textit{Stream} developed by DeepMind, which was aimed at kidney-disease prediction. Since the deep-learning version of \textit{Stream} was unable to provide interpretability of the decision-making process, it was not approved to be used in practice. Later on, the application got approval when a simple decision-tree model (which did not provide the same high accuracy as the deep-learning model) was used instead, since it provided high interpretaibility of the decision-making process.\footnote{\url{https://deepmind.com/blog/article/streams-and-ai}} 

Recent advances in the context of interpretable AI (see the excellent recent review on the topic by Fan~{\it et al.}~\cite{fan_et_al}) have brought mathematical techniques to provide interpretability properties for the training/test data set, model parameters, output, etc. This is therefore an excellent way to bring transparency to the black-box AI systems, which is such an important feature in the context of AI ethics. Here we would like to differentiate between methods that provide mere {\it explanations} regarding the AI results, and methods that yield a complete {\it interpretability} of those results. This is an important difference, and we advocate (aligned with the work by Rudin~\cite{rudin}) for methods that provide interpretability. The ideal scenario is when the AI model is interpretable from its inception~\cite{sandberg1,ml_rans}; however, in many applications, particularly dealing with deep learning, the models are already trained on extensive databases and it is costly (or impossible) to reformulate them in an interpretable framework, maintaining the accuracy. Here we will discuss some explainability methods, mention their limitations, and we will propose the use of an approach for interpretability of already-trained neural networks based on symbolic models obtained through inductive biases, recently proposed by Cranmer~{\it et al.}~\cite{cranmer_et_al} in the context of physical systems.

An example of approach providing mere explanations is the family of saliency methods, which basically identify which regions of the images or what features of the input data are more relevant to the predictions of the model. Besides the simplistic approach based on assessing the prediction changes when removing certain certain features, saliency methods typically rely on game-theory concepts such as the Shapley value~\cite{Ancona}, which quantifies the contribution of a certain feature to the predicted results. The main criticism against this type of methods~\cite{rudin} is the fact that they basically identify the parts of the input data the AI model is focused on, but they do not provide any interpretation for the actual outcome (for instance, the reason to place a certain image in one particular category). A similar criticism can be made against feature-analysis methods, which focus on the neural-network features to obtain improved explanatory power from the model~\cite{fan_et_al}. These rely on inverting-based methods that can produce synthesized images from the feature maps~\cite{Dosovitskiy}, again lacking complete interpretability regarding the final outcome of the AI model.

In our view, the method proposed by Cranmer~{\it et al.}~\cite{cranmer_et_al} is preferred, since it provides a greater interpretability power to already-trained deep-learning models. Their method is based on the following four steps: 
\begin{enumerate}[(i)]
    \item One needs to first develop a deep-learning model with a separable internal structure and an inductive bias which is relevant to the nature of the data. Note that the inductive bias constitutes the set of assumptions made on the structure of the deep-learning model to be able to generalize beyond the data seen during training. 
    \item After defining the model in (i), it is trained using the standard procedures corresponding to the chosen architecture and using the selected training database.
    \item Then, the key aspect proposed by Cranmer~{\it et al.}~\cite{cranmer_et_al} is to fit symbolic expressions to the functions composing the deep-learning model. This is done by means of a genetic algorithm which stochastically combines algebraic formulas.
    \item Finally, the internal functions are replaced by the fitted symbolic expressions.
\end{enumerate}

The power of this method lies in the possibility of analytical expressions to perform the same predictions initially carried out by the black-box deep-learning model. Here it is important to clarify that the resemblance of the predictions between the new and the original models will of course depend on the quality of the fit, but the work by Cranmer~{\it et al.}~\cite{cranmer_et_al} is very encouraging in the sense that their new symbolic-based model exhibited better generalization capabilities than the original black-box model. In their case, they illustrated the use of this approach with simple physical examples and a more complex one based on dark-matter simulation data. Their work was focused on graph neural networks (GNNs). We note that an important aspect of this process is to promote sparse models through regularization terms in the loss, effectively favoring the principle of Occam's Razor.

\begin{figure}[ht]
\centering 
\includegraphics[width=\textwidth]{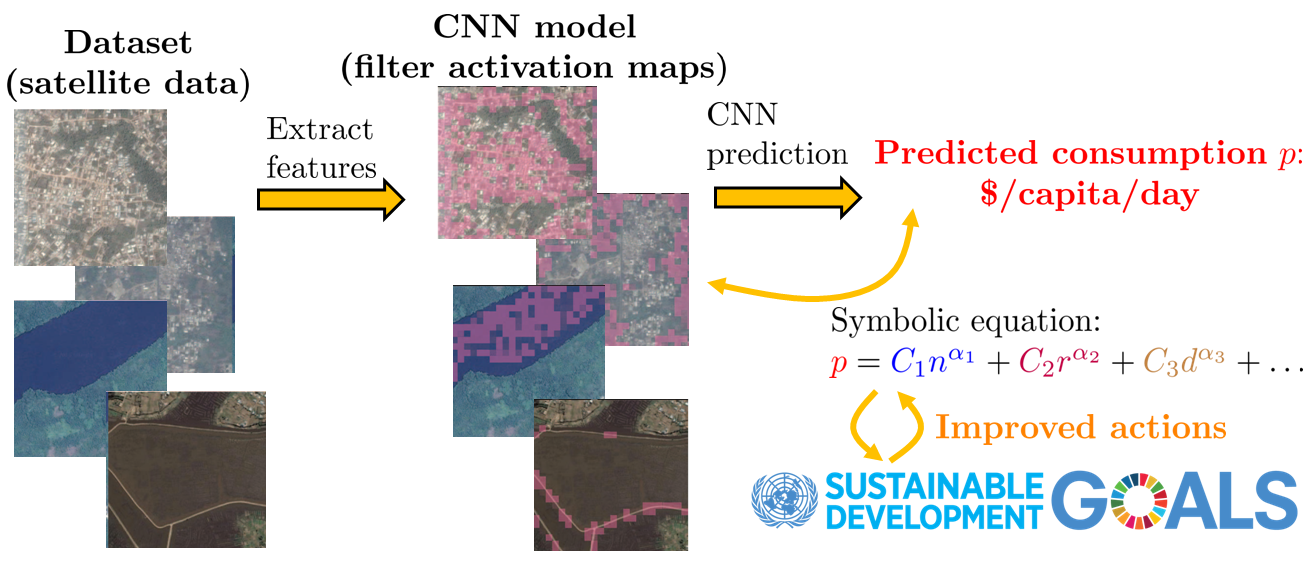}
   \caption{Schematic representation of the method proposed by Cranmer~{\it et al.}~\cite{cranmer_et_al} for adding interpretability to AI models based on symbolic equations. We use the example of poverty tracking via satellite images by Jean~{\it et al.}~\cite{jean_et_al}. Here $n$, $r$ and $d$ denote nigh-light intensity, roofing material and distance to urban areas, respectively, whereas $C_i$ and $\alpha_i$ are model constants. Note that a very simple symbolic model is provided as an example. This model will help to better understand the impact of the different variables on the poverty prediction, thus enhancing the actions to achieve SDG 1. Schematic representation adapted from Ref.~\cite{cranmer_et_al}, and panels extracted from Ref.~\cite{jean_et_al} with permission from the publisher (The American Association for the Advancement of Science).}
   \label{fig:schematic}
\end{figure}

There are many examples where black-box AI models make decisions with important implications in the individuals, and more transparency (provided through interpretability) would be highly beneficial: facial-recognition applications~\cite{naude_vinuesa}, decisions regarding assignment of loans or recruitment~\cite{vinuesa_et_al_2020}, health~\cite{Wardhana2021}, etc. Here we illustrate the use of the methodology described above with the example of tracking poverty using satellite images and CNNs~\cite{jean_et_al}. This work essentially identified features such as night-light intensity, roofing material, distance to urban areas, etc., and they predicted the average economical consumption per capita and day. The predicted-consumption values are in very good agreement with the reported ones, and their identified features can be related to around $75\%$ of the local economic results. Adding interpretability to this model would help to really understand what is the influence of each parameter on the outcome, yielding a more robust and useful tool to track poverty and coordinate actions. In fact, this type of interpretable system may be able to shed light into the dynamics of poverty, producing a deeper understanding of the current trends, and potentially being able to predict and tackle future negative developments related to SDG 1. A schematic representation of the process, as well as its implications on the SDGs, are shown in Figure~\ref{fig:schematic}. Thus, the interpretability method by Cranmer~{\it et al.}~\cite{cranmer_et_al} can provide clear understanding of the reasoning from the deep-learning models, a fact that may help both professionals and policy makers in two ways: first, being able to know how to improve the models when needed; and second, being able to design better actions aligned with the SDGs. 

To conclude, we hope that, through this Comment piece, we will be able to influence AI researchers and policymakers towards the highest benefit for society and the environment, prioritizing interpretable AI and transparency of the employed models. If such interpretable models are achieved, they would also have chances to serve in real applications, by fitting into the \textit{Trustworthy-AI} assessment guidelines provided by the \textit{European Commission}.\footnote{\url{https://www.aepd.es/sites/default/files/2019-12/ai-ethics-guidelines.pdf}}

\section*{Acknowledgements}
RV acknowledges the financial support from the Swedish Research Council (VR).

\bibliographystyle{abbrvnat}
\bibliography{interp}

\end{document}